%% file: clustering-words-by-projection-entropy.tex
\documentclass{article} 
\usepackage{nips14submit_e,times}
\usepackage[hidelinks]{hyperref}
\usepackage{url}
\usepackage{tabularx}
\usepackage{amsmath}
\usepackage{amssymb}

\usepackage{graphicx}
\usepackage{tikz}
\usetikzlibrary{decorations.pathmorphing} 
\usetikzlibrary{fit}					
\usetikzlibrary{backgrounds}	
\usetikzlibrary{matrix,positioning}

\usepackage[utf8]{inputenc}

\title{Clustering Words by Projection Entropy}

\author{
Işık Barış Fidaner\\
Computer Engineering Department\\
Boğaziçi University, Istanbul\\
\texttt{fidaner@alternatifbilisim.org}\\
\And
Ali Taylan Cemgil\\
Computer Engineering Department\\
Boğaziçi University, Istanbul\\
\texttt{taylan.cemgil@boun.edu.tr}\\
}

\nipsfinalcopy 

\begin{document}

\maketitle

\begin{abstract}
We apply entropy agglomeration (EA), a recently introduced algorithm, to cluster the words of a literary text. EA is a greedy agglomerative procedure that minimizes projection entropy (PE), a function that can quantify the segmentedness of an element set. To apply it, the text is reduced to a feature allocation, a combinatorial object to represent the word occurences in the text's paragraphs. The experiment results demonstrate that EA, despite its reduction and simplicity, is useful in capturing significant relationships among the words in the text. This procedure was implemented in Python and published as a free software: REBUS. 
\end{abstract}

\section{Introduction}\vspace*{-5pt}

Problems in natural language processing involve many difficulties due to the variety of subtleties and ambiguities in languages. On the other hand, these problems can be familiar to people in different fields, since the spoken and written languages constitute our common ground. It is especially favorable to approach these difficulties with generic statistical concepts, since these can also be utilized in other challenging fields such as bioinformatics. This paper demonstrates the use of entropy agglomeration (EA), a recently introduced algorithm [1], by clustering the words of a literary text.

By following the state-of-the-art NLP methods, we assume that \textit{words} are the basic elements of a text [2, 3]. Moreover, to approach text analysis in a simpler form, we disregard all sequential ordering, considering each paragraph as a subset of words, and the whole text as a set of paragraphs. A statistical analysis of such data would conventionally be formulated in terms of joint probabilities of word sets to \textit{co-occur} in the paragraphs. Such a probabilistic formulation can be extended to potentially infinite number of words by using Bayesian nonparametric models [4]. However, we take a different approach: we compute projection entropies (PE) of word sets, and use them in a clustering algorithm called entropy agglomeration (EA) to find the meaningful correlations among the words in the text. And we will briefly review the statistical concepts that were introduced in [1].

In the following sections, we define how the input data is represented by feature allocations, quantified by projection entropies and clustered by entropy agglomeration. We describe the experiment procedure on the way, and finally present the results, which demonstrate that our algorithm is able to capture a variety of meaningful relationships among the words of the input text. An additional discussion of projection entropy in comparison to co-occurence is included in Appendix A.\vspace{-5pt}

\section{The input text and its representation}\vspace{-5pt}

We picked \textit{Ulysses}\footnote{Full text of this novel is available at \url{http://www.gutenberg.org/ebooks/4300}} (1922) by James Joyce to be the input text. It consists of 7437 paragraphs and contains 29327 distinct words. But we only want to know which words occur in which paragraphs. This information is illustrated as a bipartite graph in Figure \ref{fig:featalloc} where word elements below are linked to the paragraphs above them. Any text with $n$ distinct words will be represented by a feature allocation defined as follows: A feature allocation of a set of elements $[n]=\{1,2,\dots,n\}$, is a multiset of blocks $F=\{B_1,\dots,B_{|F|}\}$ such that $B_i\subset[n]$ and $B_i\neq\emptyset$ for all $i\in\{1,\dots,n\}$. The blocks $B_1,\dots,B_{|F|}$ in this definition will represent the paragraphs of the input text. \input{featalloc}

See the definitions given in Figure \ref{fig:featalloc}. Entropy quantifies the segmentedness of elements with respect to the blocks of $F$. It becomes maximum at the blocks that include about half of the elements. Projection of $F$ onto $S$ restricts the scope of $F$ to this subset, functioning like a filter to focus on this particular subset. Projection entropy computes the segmentedness for a particular subset, by projecting $F$ onto it (See [1]). If a subset has low PE, we say its elements have \textit{entropic correlation}. The projection size $|PROJ(F,\{a\})|$ of an element $a$ is the number of blocks that include it. In our case, $a$ indicates a word and its projection size is the number of paragraphs it occurs in. \vspace{-5pt}

\section{Clustering the word sets}\vspace{-5pt}

\textit{Ulysses} is reduced, but the feature allocation still contains too many words. Word sets of manageable sizes are needed for analysis. Nine word sets are assembled by restricting their projection sizes (number of paragraphs they occur in) in ranges 10, 11, 12-13, 15-17, 20-25, 30-39, 40-59, 60-149 and 150-7020. Then, the full feature allocation is projected onto each of these word sets, and EA is run on each of these projected feature allocations. Here is the pseudocode for the EA algorithm:\vspace{-5pt}
\subsection*{Entropy Agglomeration Algorithm:}
{\small\begin{enumerate}
\item Initialize $\Psi \leftarrow \{\{1\},\{2\},\dots,\{n\}\}$.
\item Find the subset pair $\{S_a,S_b\}\subset \Psi$ that minimizes the entropy $H(PROJ(F,S_a\cup S_b))$.
\item Update $\Psi \leftarrow (\Psi\backslash\{S_a,S_b\})\cup\{S_a\cup S_b\}$.
\item If $|\Psi|>1$ then go to 2.
\item Generate the dendrogram of chosen pairs by plotting minimum entropies for every bifurcation.
\end{enumerate}}

EA generates a dendrogram for each word set to show the entropic correlations among its elements. Dendrograms are diagrams commonly used to display results of hierarchical clustering algorithms [1, 5]. The dendrograms generated by this procedure for \textit{Ulysses} can be examined in Appendix B.

\input{results}

Sample word pairs from EA dendrograms are shown in Table \ref{tab:results} to illustrate the variety of entropic correlations detected by the algorithm. These correlations indicate a diversity of semantic relationships: black-white, south-north are antonyms; then-now, former-latter, came-went are contraries; female-male, Eve-Adam, you-I indicate reciprocities; red-green are colors; four-five, nine-eleven are quantities; his-he, her-she, me-my, us-our, them-their, thy-thou are inflections of different pronouns; hear-heard, looking-looked, smile-smiled, pouring-poured are inflections of different verbs; ireland-irish is the inflection of a nation. Some of the other contextual correlations of expressions, things and figures are also enumerated. Entropic correlations cover an interesting range of meanings.

In conclusion, we developed in this work a text analysis tool that visualizes the entropic correlations among the words of a given text by applying entropy agglomeration to its paragraphs, and published this procedure as a free software: REBUS [6]. We demonstrated the utility of this procedure by clustering the words of a literary text using this tool.

\section*{Appendix A: On the meaning of projection entropy}

Projection entropy (PE) is a useful guiding principle in exploring significant element-wise relationships in combinatorial data. But it has a meaning that is quite different from conventional statistical methods. Therefore in this section, we would like to discuss the meaning of projection entropy in comparison to a well-known quantity, \textit{co-occurence of elements}, which is used for similar purposes in probabilistic modeling. The actualizations of these quantities' values for 2 and 3 elements are illustrated in Figure \ref{fig:entmeasure}. Firstly, as pointed out in [1], contrary to the positive sense of co-occurence as scoring the blocks where all elements co-occur; PE has a negative sense of penalizing the blocks that divide and separate them. Secondly, co-occurence is non-zero only at the blocks that include all of the elements, leaving out the blocks that exclude any of them. But PE leaves out the blocks that include all elements as well as the blocks that exclude all elements; it is non-zero only at partial inclusions: at the blocks that include some elements while excluding other elements. This makes PE a flexible quantity that can adjust to the blocks where any part of the elements overlap, whereas co-occurence is a rigid quantity that can only adjust to the blocks where all of the elements overlap.

\input{entmeasure}

Assume that we have a cluster $S$ whose elements have constant projection sizes. If these elements do not overlap at any of the blocks that include any of them, the cluster's PE will take the maximum value: the sum of projection sizes multiplied by $\frac{\log|S|}{|S|}$. If these elements overlap at all of the blocks that include them, PE will be zero by definition. Moreover, any additional overlapping among the elements will decrease the cluster's PE, if the projection sizes are kept constant. Therefore, lower PE indicates higher overlapping in the cluster, relative to its elements' projection sizes. To express this element-wise overlapping indicated by a lower PE, we say that these elements have an \textit{entropic correlation} at the blocks that include them.

To understand how PE functions in entropy agglomeration, let us examine its effect for a word pair. Assume that we have a pair from the set of words that occur exactly in 10 paragraphs. We know that (1) projection sizes for both words are 10, (2) co-occurence would count the blocks that include both of them, (3) PE would count the blocks that include one of them. For this particular case, these two quantities are directly proportional: an increase in co-occurence by $1$ would decrease PE by $\log 2$. This makes them practically equivalent. However, if there are several projection sizes like 20-25, words with larger projections can have partial occurences more frequently; co-occurence would ignore these occurences, but PE may count them to penalize the elements for occuring `unnecessarily'.

\newpage
\section*{Appendix B:\ Entropy agglomeration dendrograms for \textit{Ulysses}\footnote{\hspace{3pt} Full results of this experiment and the Python code can be found on the REBUS website [6].}}\vspace*{-7pt}
\input{appendix}

\newpage
\subsubsection*{References}

\small{
[1] Fidaner, I. B. \& Cemgil, A. T. (2013) Summary Statistics for Partitionings and Feature Allocations. In \textit{Advances in Neural Information Processing Systems}, 26.

[2] Wood, F., Archambeau, C., Gasthaus, J., James, L. F. \&  Teh, Y.W. (2011) The Sequence Memoizer. \textit{Communications of the ACM}, 54(2):91-98.

[3] Mikolov, T., Sutskever, I., Chen, K., Corrado, G. S. \& Dean, J. (2013) Distributed Representations of Words and Phrases and their Compositionality. In \textit{Advances in Neural Information Processing Systems}, 26.

[4] Teh, Y. W. (2006) A hierarchical Bayesian language model based on Pitman-Yor processes. In Proceedings of the 21st International Conference on Computational Linguistics and the 44th annual meeting of the Association for Computational Linguistics (ACL-44). Association for Computational Linguistics, Stroudsburg, PA, USA, 985-992.

[5] Eisen, M. B., Spellman, P. T., Brown, P. O., \& Botstein, D. (1998) Cluster analysis and display of genome-wide expression patterns. \textit{Proceedings of the National Academy of Sciences}, 95(25):14863-14868.

[6] Fidaner, I. B. \& Cemgil, A. T. (2014) REBUS: entropy agglomeration of text. Published under GNU General Public License. Online: \url{http://fidaner.wordpress.com/science/rebus/}

\end{document}

%% file: featalloc.tex
\begin{figure}[t]
\centering
\begin{tikzpicture}[scale=0.98]
\draw  (-5,1.5) node (v1) {} circle (0.1);
\draw  (-4,1.5) node (v3) {} circle (0.1);
\draw  (-3,1.5) node (v8) {} circle (0.1);
\draw  (-2,1.5) node (v11) {} circle (0.1);
\node at (-1,1.5) {$\dots$};
\draw  (-0.5,1.5) circle (0.1);
\draw  (-5,0) node (v2) {} circle (0.1);
\draw  (-4.5,0) node (v4) {} circle (0.1);
\draw  (-4,0) node (v5) {} circle (0.1);
\draw  (-3.5,0) node (v6) {} circle (0.1);
\draw  (-3,0) node (v7) {} circle (0.1);
\draw  (-2.5,0) node (v9) {} circle (0.1);
\draw  (-2,0) node (v10) {} circle (0.1);
\draw  (-1.5,0) node (v12) {} circle (0.1);
\node at (-1,0) {$\dots$};
\draw  (-0.5,0) circle (0.1);
\draw  (v1) edge (v2);
\draw  (v3) edge (v4);
\draw  (v3) edge (v5);
\draw  (v3) edge (v6);
\draw  (v3) edge (v7);
\draw  (v8) edge (v2);
\draw  (v8) edge (v9);
\draw  (v8) edge (v10);
\draw  (v11) edge (v7);
\draw  (v11) edge (v12);
\node at (-1,0.8) {$\dots$};
\node[align=left] at (-2.75,-0.4) {$a$\hspace{6.60pt} $b$\hspace{6.60pt} $c$\hspace{6.60pt} $d$\hspace{6.60pt} $e$\hspace{6.60pt} $f$\hspace{6.60pt} $g$\hspace{6.60pt} $h$\hspace{18pt} $x$};
\node[align=left] at (-2.60,1.8) {$B_1$\hspace{13pt} $B_2$\hspace{13pt} $B_3$\hspace{13pt} $B_4$\hspace{28pt} $B_{|F|}$};
\node at (3.90,1.5) {Entropy:\hspace{5pt} $H(F)\ =\ \sum_{i=1}^{|F|}\ \frac{|B_i|}{n} \log \frac{n}{|B_i|}$};
\node at (4.55,0.7) {Projection:\hspace{5pt} $PROJ(F,S)\ =\ \{B\cap S\}_{B\in F} \backslash \{\emptyset\}$};
\node at (3.80,-0.1) {Projection entropy:\hspace{5pt} $H(PROJ(F,S))$};
\end{tikzpicture}\vspace*{-10pt}
\caption{The text is represented by a feature allocation where each block represents a paragraph.}
\label{fig:featalloc}\vspace*{-12pt}
\end{figure}
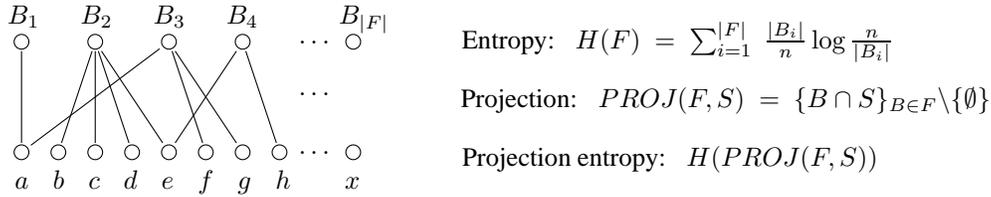

%% file: results.tex
\begin{table}[b]
\vspace{-10pt}
{\renewcommand{\tabcolsep}{3pt}
{\renewcommand{\arraystretch}{1.2}
{\centering
{\fontsize{7.85pt}{1em}\selectfont 
\begin{minipage}[t]{0.182\textwidth}
\begin{tabular}{|rcl|}
\hline
\multicolumn{3}{|c|}{\textbf{Differences:}}\\
\hline
black &–& white\\
south &–& north\\
\hline
then &–& now\\
former &–& latter\\
came &–& went\\
red &–& green\\
female &–& male\\
eve &–& adam\\
you &–& I\\
\hline
four &–& five\\
nine  &–& eleven\\
\hline
\end{tabular}
\end{minipage}%
\begin{minipage}[t]{0.195\textwidth}
\begin{tabular}{|rcl|}
\hline
\multicolumn{3}{|c|}{\textbf{Inflections:}}\\
\hline
his &–& he\\
her &–& she\\
me &–& my\\
us &–& our\\
them &–& their\\
thy &–& thou\\
\hline
hear &–& heard\\
looking &–& looked\\
smile &–& smiled\\
pouring &–& pour\\
\hline
ireland &–& irish\\
\hline
\end{tabular}
\end{minipage}%
\begin{minipage}[t]{0.202\textwidth}
\begin{tabular}{|rcl|}
\hline
\multicolumn{3}{|c|}{\textbf{Expressions:}}\\
\hline
ah &–& sure\\
ay &–& eh\\
darling &–& perfume\\
thank &–& please\\
\hline
\multicolumn{3}{|c|}{\textbf{Things:}}\\
\hline
ocean &–& level\\
waves &–& waters\\
river &–& boat\\
moon &–& stars\\
birds &–& fly\\
grass &–& fields\\
\hline
\end{tabular}
\end{minipage}%
\begin{minipage}[t]{0.205\textwidth}
\begin{tabular}{|rcl|}
\hline
window &–& seen\\
hand &–& eyes\\
face &–& head\\
cup &–& tea\\
plate &–& fork\\
food &–& eating\\
job &–& business\\
sell &–& trade\\
slice &–& quantity\\
family &–& memory\\
road &–& city\\
system &–& distance\\
\hline
\end{tabular}
\end{minipage}%
\begin{minipage}[t]{0.218\textwidth}
\begin{tabular}{|rcl|}
\hline
\multicolumn{3}{|c|}{\textbf{Figures:}}\\
\hline
girl &–& sweet\\
dame &–& joy\\
females &–& period\\
wife &–& world\\
woman &–& behind\\
gentleman &–& friend\\
gentlemen &–& friends\\
priest &–& quietly\\
reverend &–& blessed\\
christ &–& jew\\
human &–& live\\
\hline
\end{tabular}
\end{minipage}
}}}}
\caption{Sample word pairs to illustrate the entropic correlations captured by EA dendrograms}
\label{tab:results}
\end{table}

%% file: entmeasure.tex
\begin{figure}[h]
\centering
\hspace*{-5pt}
\begin{tikzpicture}[yscale=0.75,xscale=0.85]
\draw  (-6,1.8) ellipse (1 and 0.7);
\draw  (-5,1.8) ellipse (1 and 0.7);
\draw  (-2.3,3.2) ellipse (1 and 1);
\draw  (-1.3,3.2) ellipse (1 and 1);
\draw  (-1.8,2.3) ellipse (1 and 1);
\node at (-5.5,1.8) {$1$};
\node at (-6.5,1.8) {$0$};
\node at (-4.5,1.8) {$0$};
\node at (-3.9,1.2) {$0$};
\node at (-1.8,2.9) {$1$};
\node at (-1.8,3.6) {$0$};
\node at (-2.4,2.6) {$0$};
\node at (-1.2,2.6) {$0$};
\node at (-2.7,3.4) {$0$};
\node at (-0.9,3.4) {$0$};
\node at (-1.8,1.9) {$0$};
\node at (-6.2,2.8) {$a\in B$};
\node at (-4.7,2.8) {$b\in B$};
\node at (-2.5,4.5) {$a\in B$};
\node at (-1,4.5) {$b\in B$};
\node at (-1.8,1) {$c\in B$};
\node at (-0.4,2.1) {$0$};
\draw  (1.65,1.8) ellipse (1.1 and 0.7);
\draw  (2.95,1.8) ellipse (1.1 and 0.7);
\draw  (5.6,3.2) ellipse (1.2 and 1);
\draw  (7,3.2) ellipse (1.2 and 1);
\draw  (6.3,2.3) ellipse (1.2 and 1);
\node at (2.3,1.8) {$0$};
\node at (1.25,1.8) {{\tiny $\frac{1}{2}\log\frac{2}{1}$}};
\node at (3.35,1.8) {{\tiny $\frac{1}{2}\log\frac{2}{1}$}};
\node at (4,1.2) {$0$};
\node at (6.3,2.9) {$0$};
\node at (6.3,3.45) {{\tiny $\frac{2}{3}\log\frac{3}{2}$}};
\node at (5.6,2.45) {{\tiny $\frac{2}{3}\log\frac{3}{2}$}};
\node at (7,2.45) {{\tiny $\frac{2}{3}\log\frac{3}{2}$}};
\node at (5.1,3.4) {{\tiny $\frac{1}{3}\log\frac{3}{1}$}};
\node at (7.5,3.4) {{\tiny $\frac{1}{3}\log\frac{3}{1}$}};
\node at (6.3,1.9) {{\tiny $\frac{1}{3}\log\frac{3}{1}$}};
\node at (1.6,2.8) {$a\in B$};
\node at (3.1,2.8) {$b\in B$};
\node at (5.6,4.5) {$a\in B$};
\node at (7.1,4.5) {$b\in B$};
\node at (6.3,1) {$c\in B$};
\node at (7.8,2.1) {$0$};
\node at (-5.8,4.5) {Co-occurence of elements};
\node at (2.4,4.5) {Projection entropy};
\node at (-6,3.9) {$\sum_i [S\subset B_i] $};
\node at (2.4,3.9) {$H(PROJ(F,S))$};
\node at (6.3,0.3) {{\footnotesize $S=\{a,b,c\}$}};
\node at (2.5,0.3) {{\footnotesize $S=\{a,b\}$}};
\node at (-1.7,0.3) {{\footnotesize $S=\{a,b,c\}$}};
\node at (-5.5,0.3) {{\footnotesize $S=\{a,b\}$}};
\node (v1) at (0.2,4.8) {};
\node (v2) at (0.2,-0.2) {};
\draw[dashed]  (v1) edge (v2);
\end{tikzpicture}\vspace*{-10pt}
\caption{Comparing co-occurence and projection entropy in their values for 2 and 3 elements}
\label{fig:entmeasure}
\end{figure}
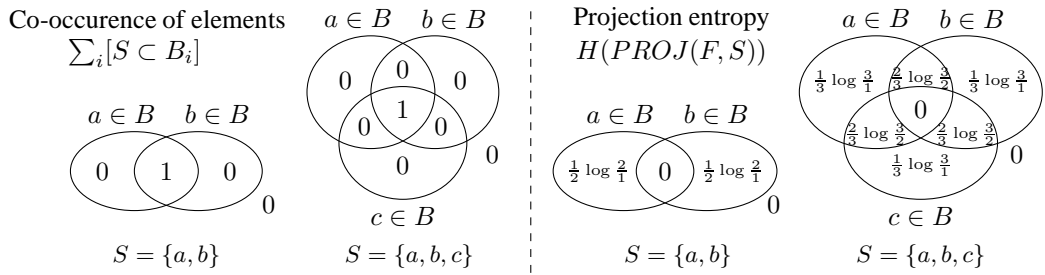

%% file: appendix.tex
{\centering
\begin{minipage}{0.2\textwidth}
\includegraphics[scale=0.27]{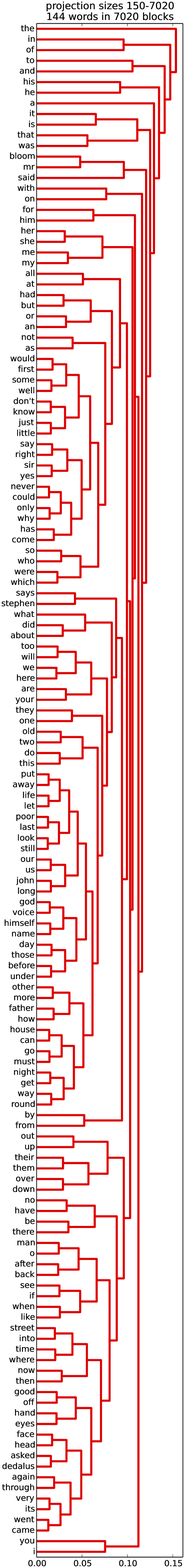}
\end{minipage}%
\begin{minipage}{0.2\textwidth}
\includegraphics[trim=0pt 1448pt 0pt 0pt,clip=true,scale=0.27]{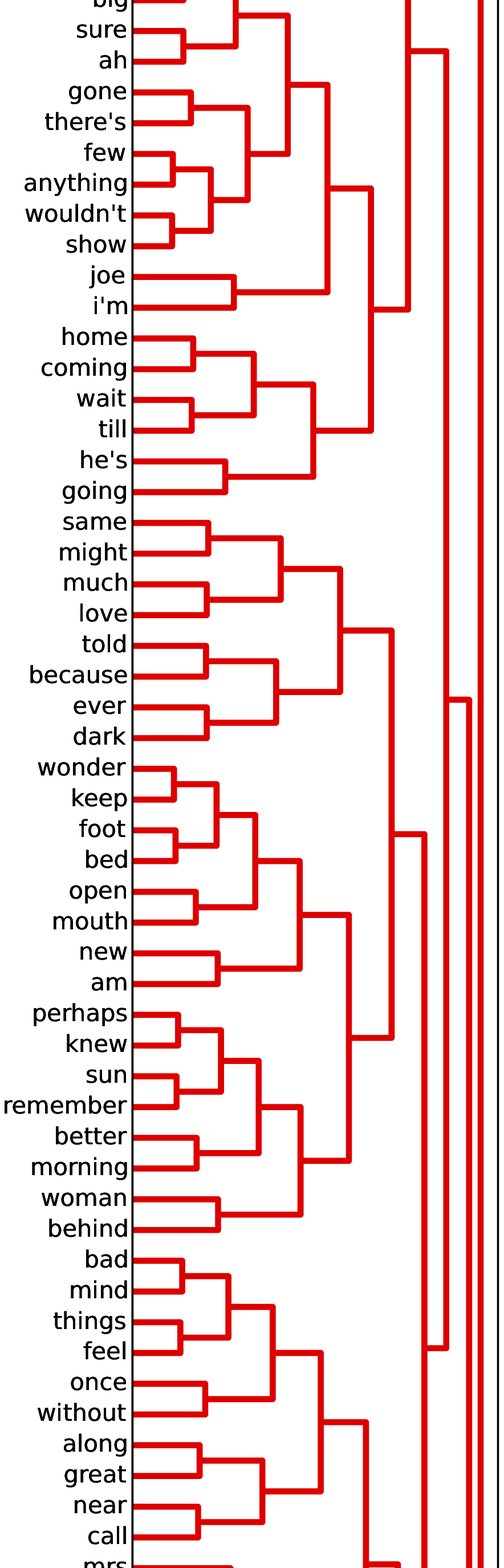}
\end{minipage}%
\begin{minipage}{0.2\textwidth}
\includegraphics[trim=0pt 0pt 0pt 2287pt,clip=true,scale=0.27]{ea_ulysses_chosen_60-149.eps}\\
\includegraphics[trim=0pt 2610pt 0pt 0pt,clip=true,scale=0.27]{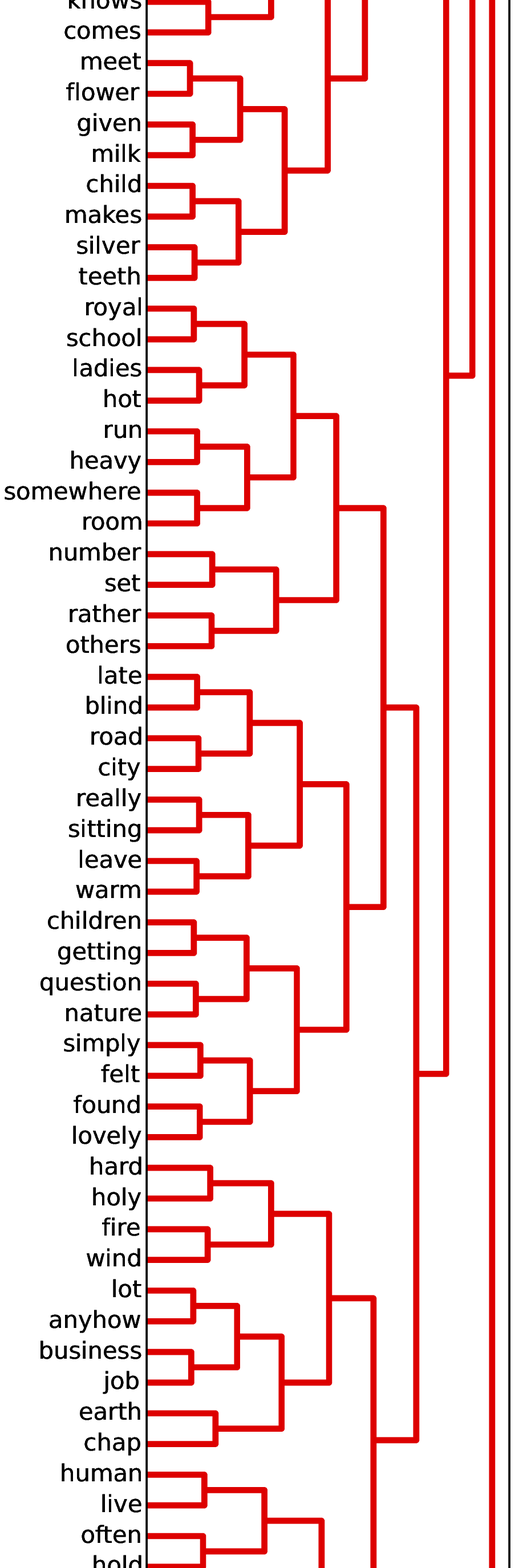}
\end{minipage}%
\begin{minipage}{0.2\textwidth}
\includegraphics[trim=0pt 355pt 0pt 805pt,clip=true,scale=0.27]{ea_ulysses_chosen_40-59.eps}
\end{minipage}%
\begin{minipage}{0.2\textwidth}
\includegraphics[trim=0pt 0pt 0pt 3062pt,clip=true,scale=0.27]{ea_ulysses_chosen_40-59.eps}\\
\includegraphics[trim=0pt 1993pt 0pt 0pt,clip=true,scale=0.27]{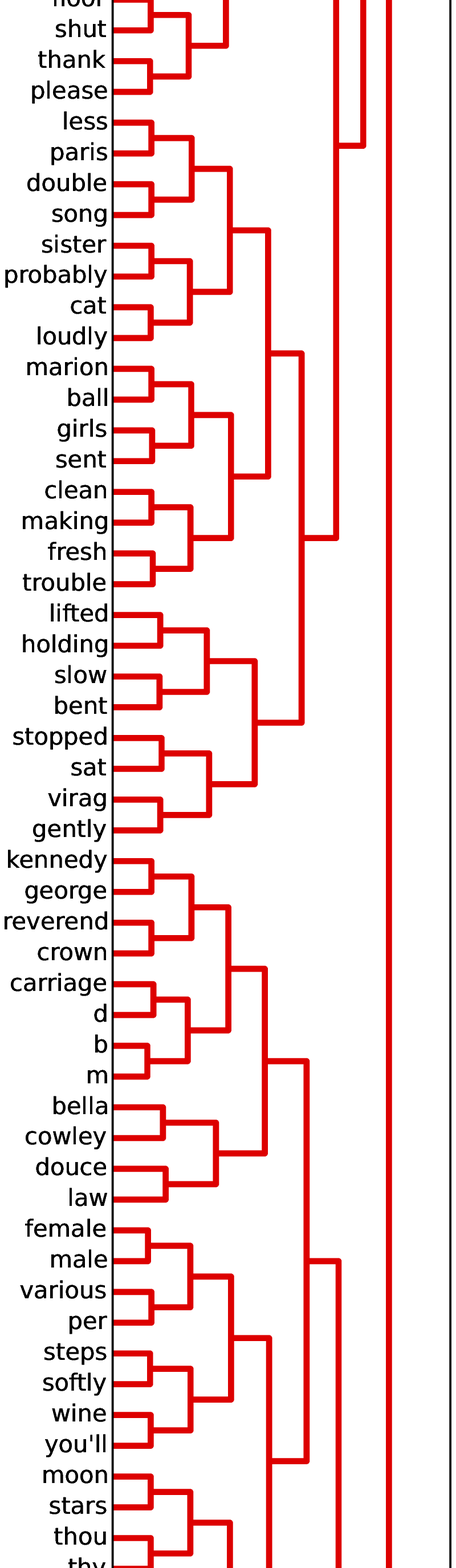}
\end{minipage}%
}